\def\year{2018}\relax
\documentclass[letterpaper]{article} 
\usepackage{aaai18}  
\usepackage{times}  
\usepackage{helvet}  
\usepackage{courier}  
\usepackage{url}  
\usepackage{graphicx}  
\usepackage{latexsym}
\usepackage{outlines}
\usepackage{amsmath}
\usepackage{subcaption}
\usepackage{color}
\begin{document}
%
\title{Fact Checking in Community Forums}
\author{Tsvetomila Mihaylova,\textsuperscript{1}
Preslav Nakov,\textsuperscript{2}
Llu\'is M\`arquez,\textsuperscript{2}
Alberto Barr\'on-Cede\~no,\textsuperscript{2}\\
{\bf \Large Mitra Mohtarami,\textsuperscript{3}
Georgi Karadzhov,\textsuperscript{1}
James Glass \textsuperscript{3}}\\
\textsuperscript{1}Sofia University ``St. Kliment Ohridski'', Sofia, Bulgaria\\
\textsuperscript{2}Qatar Computing Research Institute, Hamad bin Khalifa University, Doha, Qatar\\
\textsuperscript{3}Massachusetts Institute of Technology, Cambridge, MA, USA\\
tsvetomila.mihaylova@gmail.com,
\{pnakov, lmarquez, albarron\}@hbku.edu.qa, \\
mitra@csail.mit.edu, georgi.m.karadjov@gmail.com,
glass@mit.edu
}


\newcommand{\Ni}{({\em i})~}
\newcommand{\Nii}{({\em ii})~}
\newcommand{\Niii}{({\em iii})~}
\newcommand{\Niv}{({\em iv})~}
\newcommand{\Nv}{({\em v})~}
\newcommand{\Na}{({\em a})~}
\newcommand{\Nb}{({\em b})~}
\newcommand{\Nc}{({\em c})~}

\newcommand{\good}{\textsc{Good}\,}
\newcommand{\bad}{\textsc{Bad}\,}
\newcommand{\potential}{\textsc{Potentially Useful}\,}

\newcommand{\userCategories}{User posts categories}
\newcommand{\userQuality}{User posts quality}
\newcommand{\userActivity}{User activity}
\newcommand{\contentLingf}{Linguistic bias, subjectivity and sentiment}
\newcommand{\contentLingfExtended}{Linguistic bias, subjectivity and sentiment}
\newcommand{\contentCredibility}{Credibility}
\newcommand{\contentSentiment}{Sentiment}
\newcommand{\contentVecGoogle}{Embeddings$_{Google}$}
\newcommand{\contentVecQL}{Embeddings$_{QL}$}
\newcommand{\forumRankSameUser}{Support from the current thread}
\newcommand{\forumVecCosinesThread}{Answer--thread similarity}
\newcommand{\forumIRQLsplit}{Support from all of Qatar Living}
\newcommand{\forumHighQualPosts}{Support from high-quality posts in Qatar Living}
\newcommand{\extIRFBingWebpageQataronlySplitted}{Support from the Web}

\newcommand{\mycite}[1]{\citeauthor{#1}~\shortcite{#1}}

\newcommand{\blue}{\textcolor{blue}}
\newcommand{\red}{\textcolor{red}} 
\newcommand{\abc}[1]{\textcolor{magenta}{#1}}
\newcommand{\green}{\textcolor{green}} 
\newcommand{\rel}{\nobreak{\sc Rel}}
\newcommand{\lluis}[1]{\textcolor{blue}{\noindent LM: #1}}


\maketitle
\begin{abstract}
 
Community Question Answering (cQA) forums are very popular nowadays, as they represent effective means for communities around particular topics to share information.
Unfortunately, this information is not always factual.
Thus, here we explore a new dimension in the context of cQA, which has been ignored so far: checking the veracity of answers to particular questions in cQA forums.
As this is a new problem, we create a specialized dataset for it.
We further propose a novel multi-faceted model,
which captures information from the answer content (\emph{what is said and how}), from the author profile (\emph{who says it}), from the rest of the community forum (\emph{where it is said}), and from external authoritative sources of information (\emph{external support}). Evaluation results show a MAP value of 86.54, which is 21 points absolute above the baseline.
\end{abstract}

\section{Introduction}
Community Question Answering (cQA) forums such as StackOverflow, Yahoo! Answers, and Quora are very popular nowadays, as they represent effective means for communities around particular topics to share information
and to collectively satisfy their information needs.  
However, the information being shared is not always factual. There are multiple factors explaining the presence of incorrect answers in cQA forums, e.g., misunderstanding, ignorance, or maliciousness of the responder.
This is exacerbated by the fact that most cQA forums are barely moderated and lack systematic quality control.
Moreover, in our dynamic world of today, truth is often time-sensitive: what was true yesterday may become false today. 

We explore a new dimension in the context of cQA: checking the veracity of answers to a given question. This aspect has been ignored so far, e.g., in recent cQA tasks at NTCIR and SemEval~\cite{Ishikawa:10,nakov-EtAl:2015:SemEval,nakov-EtAl:2016:SemEval,nakov-EtAl:2017:SemEval}, where an answer is considered as \good\ if it tries to address the question, irrespective of its veracity.
Yet, veracity is an important aspect, as high-quality automatic fact checking can offer better user experience for cQA systems.
For instance, the user could be presented with 
veracity scores, where low scores would warn him/her not to completely trust the answer or to double-check it.

\begin{figure}
\centering
\small
\begin{description}\footnotesize
\item[$q$:] I have heard its not possible to extend visit visa more than 6 months? Can U please answer me.. Thankzzz...
\item[$a_1$:] Maximum period is 9 Months....
\item[$a_2$:] 6 months maximum
\item[$a_3$:] This has been anwered in QL so many times. Please do search for information regarding this. BTW answer is 6 months.
\end{description}
\caption{\label{fig:example}Example from the Qatar Living forum.}
\label{fig:example1}
\end{figure}

\noindent Figure~\ref{fig:example1} presents an excerpt of an example from the Qatar Living forum, with one question and three answers selected from a longer thread. According to SemEval-2017 Task~3~\cite{nakov-EtAl:2017:SemEval}, all three answers are good since they address the question $q$. Nevertheless, $a_1$ contains false information, while $a_2$ and $a_3$ are true,\footnote{One could also guess that $a_2$ and $a_3$ are more likely to be true from the fact that the \emph{6 months} answer appears many times in the thread, as well as in other threads. While these observations serve the basis for useful features for classification, the real verification for a gold standard annotation requires finding support from a credible external source.} as can be checked on an official governmental website.\footnote{\url{https://www.moi.gov.qa/site/english/departments/PassportDept/news/2011/01/03/23385.html}}

Determining the veracity of a claim is a very difficult problem, and solving it in full would require language understanding and inference, integration of several sources of information, and world knowledge, among other things. Here, we approach it as a supervised classification task, and we propose a novel model based on multi-faceted modeling of the facts, which integrates knowledge from several complementary sources, such as the answer content (\emph{what is said and how}), the author profile (\emph{who says it}), the rest of the community forum (\emph{where it is said}), and external authoritative sources of information (\emph{external support}).

The main contributions of this paper are as follows:
\Ni first, we are the first to study factuality in cQA, and we create a new high-quality dataset ---CQA-QL-2016-fact---, which we release to the research community;%
\footnote{The dataset and the source code are available online at \url{https://github.com/qcri/QLFactChecking}}
to the best of our knowledge, this is the first publicly-available dataset specifically targeting factuality in a cQA setting;
\Nii we approach the problem of fact-checking using a multi-faceted model based on a rich input representation, including new features that have not been compared in such a configuration before; 
\Niii this rich representation allows us to obtain strong results that are applicable to supporting in practice the application scenario outlined above; and  
\Niv we perform a qualitative analysis of what works well and what does not.

\section{Related Work}

To the best of our knowledge, no previous work has targeted fact-checking of answers in the context of community Question Answering.
Yet, there has been work on \emph{credibility} assessment in cQA~\cite{RANLP2017:credibility:trolls}. 
However, \emph{credibility} is different from \emph{veracity} (our focus here) as it is a subjective perception about whether a statement is credible, rather than verifying it as true/false as a matter of fact. 

In the context of general QA, there has been work on \emph{credibility} assessment which has been only modeled at the feature level, with the goal of improving \good\ answer identification. 
For example, \mycite{Jurczyk:2007:DAQ:1321440.1321575} modeled author authority using link analysis, while~\mycite{Agichtein:2008:FHC:1341531.1341557} used  
PageRank and HITS in addition to intrinsic content quality (e.g., punctuation and typos, syntactic and semantic complexity, and grammaticality), and usage analysis (e.g., number of clicks and dwell time).

In~\cite{lita2005qualitative} the focus was on source credibility, sentiment analysis, and answer contradiction compared to other answers, while in~\cite{Su-EtAl:2010:PACLIC2010} the emphasis was on verbs and adjectives that cast doubt. Other authors used language modeling to validate the reliability of an answer's source~\cite{banerjee-han:2009:NAACLHLT09-Short} or focused on non-textual features such as click counts, answer activity level, and copy counts \cite{Jeon:2006:FPQ:1148170.1148212}. 
There has been also work on curating social media content using syntactic, semantic, and social signals \cite{Pelleg:2016:CEI:2818048.2820022}. Unlike this research, we 
\Ni target factuality rather than credibility, 
\Nii address it as a task in its own right, 
\Niii use a specialized dataset, and 
\Niv use a much richer text representation.

Information credibility, fact-checking and rumor detection have been also studied in the area of social computing. \mycite{Castillo:2011:ICT:1963405.1963500} used user reputation, author writing style, and various time-based features.
\mycite{Canini:2011} analyzed the interaction of content and social network structure and 
\mycite{Morris:2012:TBU:2145204.2145274} and \mycite{PlosONE:2016} studied how people handle rumors in social media.
\mycite{lukasik-cohn-bontcheva:2015:ACL-IJCNLP} used temporal patterns to detect rumors and to predict their frequency.
\mycite{Ma:2015:DRU} further used recurrent neural networks,
and~\mycite{PlosONE:2016} focused on conversational threads.
Other authors have gone beyond social media and have been querying the Web to gather support for accepting or refuting a claim~\cite{popat2016credibility}.
Finally, there has been also work on studying credibility, trust, and expertise in
news communities~\cite{mukherjee2015leveraging}.
However, none of this work was about QA or cQA.

\section{CQA-QL-2016-fact: A Dataset for Fact Checking in cQA}
\label{sec:corpus}

As we have a new problem ---fact-checking of answers in the context of cQA---, for which no dataset exists, we had to create our own one. We chose to augment with factuality annotations a pre-existing dataset for cQA, which allows us to stress the difference between (a)~distinguishing a \good\ vs.\ a \bad\ answer, and (b)~distinguishing between a factually-true vs.\ a factually-false one.
In particular, we added annotations for factuality to the CQA-QL-2016 dataset from SemEval-2016 Task 3\footnote{\url{http://alt.qcri.org/semeval2016/task3/}} on Community Question Answering.

In CQA-QA-2016, the data is organized in question--answer threads from the Qatar Living forum.%
\footnote{\url{http://www.qatarliving.com/forum}}
Each question has a subject, a body, and meta information: ID, category (e.g., \emph{Computers and Internet}, \emph{Education}, and \emph{Moving to Qatar}), date and time of posting, user name and ID. 

We selected for annotation only the factual questions such as ``\emph{What is Ooredoo customer service number?}'' In particular, we filtered out all
(\emph{i})~socializing, e.g., ``\emph{What was your first car?}'', (\emph{ii})~requests for opinion/advice/guidance, e.g.,~``\emph{Which is the best bank around??}'', and (\emph{iii})~questions containing multiple sub-questions, e.g., ``\emph{Is there a land route from Doha to Abudhabi. If yes; how is the road and how long is the journey?}''

Next, we annotated for veracity the answers to the questions that we retained in the previous step.
In CQA-QA-2016, each answer has a subject, a body, meta information (answer ID, user name and ID), and a judgment about how well it answers the question of its thread: \good\  vs. \bad\  vs. \potential. 
We only annotated the \good\  answers, using the following labels:

\textsc{Factual - True}: 
The answer is True and this can be manually verified using a trusted external resource.
(\emph{Q: ``I wanted to know if there were any specific shots and vaccinations I should get before coming over [to Doha].''; A: ``Yes there are; though it varies depending on which country you come from. In the UK; the doctor has a list of all countries and the vaccinations needed for each.''}).\footnote{This can be verified: \url{https://wwwnc.cdc.gov/travel/destinations/traveler/none/qatar}}

\textsc{Factual - False}:
The answer gives a factual response, but it is false.
(\emph{Q: ``Can I bring my pitbulls to Qatar?'', A:~``Yes you can bring it but be careful this kind of dog is very dangerous.''}).\footnote{The answer is not true because pitbulls are included in the list of breeds that are banned in Qatar: \url{http://canvethospital.com/pet-relocation/banned-dog-breed-list-qatar-2015/}}

\textsc{Factual - Partially True}:
We could only verify part of the answer.
(\emph{Q: ``I will be relocating from the UK to Qatar [...] is there a league or TT clubs / nights in Doha?'', A:~``Visit Qatar Bowling Center during thursday and friday and you'll find people playing TT there.''}).\footnote{The place has table tennis, but we do not know on which days: \url{https://www.qatarbowlingfederation.com/bowling-center/}}

\textsc{Factual - Conditionally True}:
The answer is True in some cases, and False in others, depending on some conditions that the answer does not mention.
(\emph{Q: ``My wife does not have NOC from Qatar Airways; but we are married now so can i bring her legally on my family visa as her husband?'', A: ``Yes you can.''}).\footnote{This answer can be true, but this depends upon some conditions: \url{http://www.onlineqatar.com/info/dependent-family-visa.aspx}}

\textsc{Factual - Responder Unsure}:
The person giving the answer is not sure about the veracity of his/her statement.
(e.g., ``\emph{Possible only if government employed. That's what I heard.}'')

\textsc{NonFactual}:  The answer is not factual.
It could be an opinion, advice, etc.\ that cannot be verified.
(e.g., ``\emph{Its better to buy a new one.}'')

We further discarded answers whose factuality was very time-sensitive (e.g., ``\emph{It is Friday tomorrow.}'', ``\emph{It was raining last week.}'')\footnote{Arguably, many answers are somewhat time sensitive, e.g., ``\emph{There is an IKEA in Doha.}'' is true only after IKEA opened, but not before that. In such cases, we just used the present situation (Summer 2017) as a point of reference.},
or for which the annotators were unsure. 

We considered all questions from the \textsc{Dev} and the \textsc{Test} partitions of the CQA-QA-2016 dataset.
We targeted very high quality, and thus we did not use crowdsourcing for the annotation, as pilot annotations showed that the task was very difficult and that it was not possible to guarantee that \textit{Turkers} would do all the necessary verification; e.g.,~gather evidence from trusted sources. Instead, all examples were first annotated independently by four annotators,
and then they discussed \emph{each example} in detail to come up with a final consensus label.
We ended up with 249 \good answers\footnote{This is comparable in size to other fact-checking datasets, e.g., \mycite{Ma:2015:DRU} experimented with 226 rumors, and \mycite{popat2016credibility} used 100 Wiki hoaxes.} to 71 different questions, which we annotated for factuality: 128 \textsc{Positive} and 121 \textsc{Negative} examples. See  Table~\ref{table:comment-labels-distribution} for more detail.

\begin{table}[t]\footnotesize
\centering
\begin{tabular}{llccc}
& \textbf{Label} & \textbf{Answers} \\
\hline
$+$ & \textsc{Factual - True} 			& 128 \\
\hline
$-$ & \textsc{Factual - False} 			& \,\,\,22 \\
$-$ & \textsc{Factual - Partially True} 	& \,\,\,38 \\
$-$ & \textsc{Factual - Conditionally True}	& \,\,\,16 \\
$-$ & \textsc{Factual - Responder Unsure}	& \,\,\,26 \\
$-$ & \textsc{NonFactual}			& \,\,\,19 \\
\hline
& $+$ \bf \textsc{Positive} & \bf 128\\
& $-$ \bf \textsc{Negative} & \bf 121\\
& \bf TOTAL & \bf 249 \\
\hline
\end{tabular}
\caption{Distribution of the answer labels in the CQA-QL-2016-fact dataset.}
\label{table:comment-labels-distribution}
\end{table}

\section{Modeling Facts}

We use a multi-faceted model, based on a rich input representation that models
\Ni the user profile, \Nii the language used in the answer, \Niii the context in which the answer is located, and \Niv external sources of information.

\subsection{User Profile Features (\emph{who says it})}
\label{subsec-userprofile}
These are features characterizing the user who posted the answer, previously proposed for predicting credibility in cQA \cite{RANLP2017:credibility:trolls}.

\paragraph{\userCategories} (\emph{396 individual features}) We count the answers a user has posted in each of the 197 categories in Qatar Living. We have each feature twice: once raw and once normalized by the total number of answers $N$ the user has posted. We further use as features this $N$, and the number of distinct categories the user has posted in.

\paragraph{\userQuality} (\emph{13 features}) 
We first use the CQA-QA-2016 data to train a \good\ vs. \bad\ answer classifier, as described by~\mycite{barroncedeno-EtAl:2015:ACL-IJCNLP}.
We then run this classifier (which has 80+\% accuracy) on the entire unannotated Qatar Living database (2M answers, provided by the SemEval-2016 Task 3 organizers) and we aggregate its predictions to build a user profile:
number of \good/\bad\ answers, total number of answers, percentage of \good/\bad\ answers, sum of the classifier's probabilities for \good/\bad\ answers, total sum of the classifier's probabilities over all answers, average score for the probability of \good/\bad\ answers, and highest absolute score for the probability of a \good/\bad\ answer.

\paragraph{\userActivity} (\emph{19 features}) 
These features describe the overall activity of the user. 
We include the number of answers posted, number of distinct questions answered, number of questions asked, number of posts in the \emph{Jobs} and in the \emph{Classifieds} sections, number of days since registering in the forum, and number of active days. We also have features modeling the number of answers posted during working hours (7:00-15:00h)\footnote{This is forum time, i.e.,~local Qatar time.}, after work, at night, early in the morning, and before noon. We also model the day of posting: during a working day vs.\ during the weekend. Finally, we track the number of answers posted among the first $k$ in a question--answer thread, for $k \in \{1,3,5,10,20\}$.

\subsection{Answer Content (\em{how it is said})} 
\label{subsec-answer}

These features model what the answer says, and how. Such features were previously used by~\mycite{RANLP2017:debates}.

\subsubsection{\contentLingf}\label{LinguisticAnalysis}

Forum users (consciously or not), often put linguistic markers in their answers, which can signal the degree of the user's certainty in the veracity of what they say. Table~\ref{table:bias_types} lists some categories of such markers, together with examples.

We use linguistic markers such as \textit{factives} from~\cite{hooper1974assertive}, \textit{assertives} from~\cite{hooper1974assertive}, \textit{implicatives} from~\cite{karttunen1971Implicatives},  \textit{hedges} from~\cite{hyland2005metadiscourse}, \textit{Wiki-bias} terms from~\cite{Recasens:ACL:13}, \textit{subjectivity} cues from~\cite{Riloff:2003:LEP:1119355.1119369}, and 
\textit{sentiment} cues from~\cite{Liu:2005:OOA:1060745.1060797}.\footnote{Most of these bias cues can be found at \url{https://people.mpi-sws.org/~cristian/Biased_language.html}} 


\emph{Factives} (\emph{1 feature}) 
are verbs that imply the veracity of their complement clause. For example, in \textit{E1} below, \textit{know} suggests that ``they will open a second school \dots'' and ``they provide a qualified french education \dots'' are factually true statements.

\begin{description}
\item[\it{E1:}] 
\begin{itemize}\footnotesize
\item[Q:] {What do you recommend as a French school; Lycee Voltaire or Lycee Bonaparte?}
\item[A:] {... About Voltaire; I \textit{\textbf{know}} that they \textit{\textbf{will}} open a second school; and they are a \textit{\textbf{nice}} french school... I \textit{\textbf{know}} that they \textit{\textbf{provide}} a \textit{\textbf{qualified}} french education and add with that the history and arabic language to be adapted to the qatar. I \textit{\textbf{think}} that's an \textit{\textbf{interesting}} addition.}
\end{itemize}
\end{description}

\begin{table}[t]
\footnotesize
\centering
\begin{tabular}{ll}
\textbf{Bias Type} & \textbf{Sample Cues} \\
\hline
Factives & realize, know, discover, learn\\ 
Implicatives & cause, manage, hesitate, neglect \\ 
Assertives & think, believe, imagine, guarantee\\ 
Hedges & approximately, estimate, essentially\\ 
Report-verbs & argue, admit, confirm, express\\ 
Wiki-bias & capture, create, demand, follow\\ 
\hline
Modals & can, must, will, shall\\
Negations & neither, without, against, never, none\\
\hline
Strong-subj & admire, afraid, agreeably, apologist\\
Weak-subj & abandon, adaptive, champ, consume\\ 
Positives & accurate, achievements, affirm\\ 
Negatives & abnormal, bankrupt, cheat, conflicts\\
\hline
\end{tabular}
\caption{Some cues for various bias types.}
\label{table:bias_types}
\end{table}

\emph{Assertives} (\emph{1 feature}) 
are verbs that imply the veracity of their complement clause with some level of certainty. For example, in \textit{E1}, \textit{think} indicates some uncertainty, while verbs like \textit{claim} cast doubt on the certainty of their complement clause.

\emph{Implicatives} (\emph{1 feature}) 
imply the (un)truthfulness of their complement clause, e.g., \textit{decline} and \textit{succeed}.

\emph{Hedges} (\emph{1 feature}) 
reduce 
commitment to the truth, e.g.,~\textit{may} and \textit{possibly}.

\emph{Reporting verbs} (\emph{1 feature})
are used to report a statement from a source, e.g., \textit{argue} and \textit{express}.

\emph{Wiki-bias} (\emph{1 feature}) 
This feature involves bias cues extracted from the NPOV Wikipedia corpus \cite{Recasens:ACL:13}, e.g.,~\textit{provide} (in \textit{E1}), and controversial words such as \textit{abortion} and \textit{execute}.

\emph{Modals} (\emph{1 feature})
can change certainty (e.g.,~\textit{will} or \textit{can}), make an offer (e.g.,~\textit{shall}), ask permission (e.g.,~\textit{may}), or express an obligation or necessity (e.g., \textit{must}).

\emph{Negation} (\emph{1 feature})
cues are used to deny or make negative statements, e.g.,~\textit{no}, \textit{never}.

\emph{Subjectivity} (\emph{2 features}) 
is used when a question is answered with personal opinions and feelings. There are two types of subjectivity cues: \textit{strong} and \textit{weak}. For example, in \textit{E1}, \textit{nice} and \textit{interesting} are \textit{strong} subjectivity cues, while \textit{qualified} is a \textit{weak} one.

\emph{Sentiment cues} (\emph{2 features}) 
We use \textit{positive} and \textit{negative}
sentiment cues 
to model the attitude, thought, and emotions of the person answering. For example, in \textit{E1}, \textit{nice}, \textit{interesting} and \textit{qualified} are positive cues.

\noindent The above cues are about single words.
We further generate multi-word cues by combining \emph{implicative}, \emph{assertive}, \emph{factive} and \emph{report} verbs with first person pronouns (\emph{I/we}), \emph{modals} and strong subjective \emph{adverbs}, e.g.,  \emph{I/we+verb} (e.g. ``I believe''), \emph{I/we+adverb+verb} (e.g., ``I certainly know''), \emph{I/we+modal+verb} (e.g., ``we could figure out'') and \emph{I/we+modal+adverb+verb} (e.g., ``we can obviously see'').

Finally, we compute a feature vector for an answer using these cues according to Equation~\eqref{Ling-equation}, where for each bias type $B_i$ and answer $A_j$, the frequency of the cues for $B_i$ in $A_j$ is normalized by the total number of words in $A_j$:
\begin{equation}\label{Ling-equation}\footnotesize
B_i(A_j) = \dfrac{\sum\limits_{cue \in B_i} {count(cue, A_j)}}{\sum\limits_{w_k \in A_j} {count(w_k, A_j)}}
\end{equation}

\subsubsection{Quantitative Analysis:}

\textbf{\contentCredibility} (\emph{31 features})
We use features 
that have been previously proposed for credibility detection \cite{Castillo:2011:ICT:1963405.1963500}:
number of URLs/images/emails/phone numbers;
number of tokens/sentences;
average number of tokens;
number of 1st/2nd/3rd person pronouns;
number of positive/negative smileys;
number of single/double/triple exclamation/in\-terrogation symbols.
To this set, we further add 
number of interrogative sentences;
number of nouns/verbs/adjectives/adverbs/pronouns;
and number of words not in word2vec's Google News vocabulary (such OOV words could signal slang, foreign language, etc.).

\subsubsection{Semantic Analysis:}

\textbf{\contentVecGoogle}
(\emph{300 features}) We use the pre-trained, 300-dimensional embedding vectors that~\mycite{mikolov-yih-zweig:2013:NAACL-HLT} trained on 100 billion words from Google News. We compute a vector representation for an answer by simply averaging the embeddings of the words it contains.

\subsubsection{Semantic Analysis:}

\textbf{\contentVecQL} (\emph{100 features})
We also use 100-dimensional word embeddings from~\cite{SemEval2016:task3:SemanticZ}, trained on all Qatar Living.

\subsection{External Evidence (\emph{external support})}
\label{subsec-external}

Following~\mycite{RANLP2017:factchecking:external}, we tried to verify whether an answer's claim is true by searching for support on the Web. We started with the concatenation of an answer to its question. Then, following~\mycite{potthast2013overview}, we extracted nouns, verbs and adjectives, sorted by TF-IDF (IDF computed on Qatar Living). We further extracted and added the named entities from the text and we generated a query of 5-10 words. If we did not obtain ten results, we dropped some terms from the query and we tried again. 

\paragraph{\extIRFBingWebpageQataronlySplitted} (\emph{180~features}): We automatically queried Bing\footnote{We also experimented with Google and the aggregation of Bing and Google, with slightly worse results.} and extracted features from the resulting webpages, excluding those that are not related to Qatar. 
In particular, we calculated similarities: \Ni cosine with TF-IDF weighting, \Nii cosine using Qatar Living embeddings, and \Niii containment~\cite{lyon2001detecting}. 

\noindent We calculated these similarities between, on the one hand, \Ni the question or \Nii the answer or \Niii the question-answer pair, vs.\ on the other hand, (a)~the snippets or (b)~the web pages. 
In order to calculate the similarity against a webpage, we first converted that webpage into a list of rolling sentence triplets. Then we calculated the score of the Q/A/Q-A vs.\ this triplet, and finally we took the average and also the maximum similarity over these triplets.
Now, as we had up to ten Web results, we further took the maximum and the average over all the above features over the returned Qatar-related pages.

We created three copies of each feature, depending on whether it came \Ni from a reputed source (e.g., news, government websites, official sites of companies), \Nii from a forum-type site (forums, reviews, social media), or \Niii from some other type of websites.

\subsection{Intra-forum Evidence \emph(\emph{where it is said})}
\label{subsec-context}

\subsubsection{Intra-thread Analysis:}

\textbf{\forumRankSameUser} (\emph{3 features})
:We use the cosine similarity between an answer- and a thread-vector of all \good\ answers using 
\contentVecGoogle\  and \contentVecQL.
The idea is that if an answer is similar to other answers in the thread, it is more likely to be true. 
To this, we add a feature for the reciprocal rank of the answer in the thread, assuming that more recent answers are more likely to be up-to-date and factually true.

\subsubsection{Forum-Level Evidence:}

\textbf{\forumIRQLsplit} (\emph{60 features})
We further collect supporting evidence from all threads in the Qatar Living forum. We use a search engine as for the external evidence features above, but this time we limit the search to the Qatar Living forum only.

\subsubsection{Forum-Level Evidence:}

\textbf{\forumHighQualPosts}
\label{Inter-thread-evidence} (\emph{10 features})
Among the 60,000 active users of the Qatar Living forum, there is a community of 38 trusted users who have written 5,230 high-quality articles on topics that attract a lot of interest, e.g., visas, work legislation, etc.
We try to verify the answers against these high-quality posts.
\Ni Since an answer can combine both relevant and irrelevant information with respect to its question, we first generate a query as explained above for each Q\&A.
\Nii We then compute cosines between the query and the sentences in the high-quality posts, and we select the $k$-best matches. 
\Niii Finally, we compute textual entailment scores \cite{Kouylekov:2010} for the answer given the $k$-best matches, which we then use as features.

\begin{table*}[t]
\footnotesize
\centering
\begin{tabular}{clccccc}
\textbf{Rank} & \textbf{Feature Group / System} & \textbf{Acc} & \textbf{P}  & \textbf{R}  & \textbf{F$_1$}  & \textbf{MAP}   \\
\hline
\\
\multicolumn{7}{c}{\textbf{External Evidence}} \\
\hline
2 & \extIRFBingWebpageQataronlySplitted & 63.45 & 59.59 & 89.84 & 71.65 & 67.71 \\
\hline
\\
\multicolumn{7}{c}{\textbf{Intra-Forum Evidence}} \\
\hline
1 & \forumIRQLsplit & 65.46 & 66.41 & 66.41 & 66.41 & 83.97 \\
4 & \forumHighQualPosts & 60.24 & 61.60 & 60.16 & 60.87 & 74.50 \\
7 & \forumRankSameUser		& 53.41 & 53.53 & 71.09 & 61.07 & 64.15 \\

\hline
\\
\multicolumn{7}{c}{\textbf{Answer Content}} \\
\hline
3 & \contentLingfExtended & 60.64 & 60.42 & 67.97 & 63.97 & 78.81 \\
5 & \contentVecQL	& 59.44 & 59.71 & 64.84 & 62.17 & 75.63 \\
6 & \contentCredibility & 56.23 & 56.21 & 67.19 & 61.21 & 64.92 \\
8 & \contentVecGoogle	& 52.61 & 53.62 & 57.81 & 55.64 & 69.23 \\

\hline
\\
\multicolumn{7}{c}{\textbf{User Profile}} \\
\hline
9 & \userActivity	& 42.57 & 46.67 & 82.03 & 59.49 & 69.04 \\
10 & \userCategories	& 42.57 & 46.67 & 82.03 & 59.49 & 68.50 \\
11 & \userQuality	& 28.92 & 31.01 & 31.25 & 31.13 & 67.43 \\
\hline
\\
\multicolumn{7}{c}{\textbf{Ensemble Systems}} \\
\hline
 & Optimizing for Accuracy & \textbf{72.29} & 70.63 & 78.91 & \textbf{74.54} & 74.32 \\
 & Optimizing for MAP & 69.88	& \bf 70.87 & 70.31 & 70.59 & \textbf{86.54}\\
\hline
\\
\multicolumn{7}{c}{\textbf{Baselines}} \\
\hline
 & \contentCredibility\  \cite{Castillo:2011:ICT:1963405.1963500} & 56.23 & 56.21 & 67.19 & 61.21 & 64.92 \\
 & All \textsc{Positive} (majority class) & 51.41 & 51.41 & 100.00 & 67.91 & --- \\
 & Thread order (chronological) & --- & --- & --- & --- & 63.75 \\
\hline
\end{tabular}
\caption{Experimental results for different feature groups as well as for ensemble systems and for some baselines. The first column shows the rank of each feature group, based on accuracy. The following columns describe the feature group and report accuracy (Acc), precision (P), recall (R), F$_1$, and mean-average precision (MAP).}
\label{table:all-feature-groups-all-results}
\end{table*}

\section{Evaluation and Results}

\subsection{Settings}

We train an SVM classifier~\cite{Joachims:1999:MLS:299094.299104}
on the 249 examples as described above,
where each example is one question--answer pair.
For the evaluation, we use leave-one-thread-out cross validation,
where each time we exclude and use for testing one of the 71 questions together with all its answers. 
We do so in order to respect the structure of the threads when splitting the data.
We report Accuracy, Precision, Recall, and F$_1$ for the classification setting. We also calculate Mean Average Precision (MAP).

\subsection{Results}

Table~\ref{table:all-feature-groups-all-results} shows results for each of the above-described feature groups, further grouped by type of evidence ---external, internal, answer-based, or user-related---, as well as for ensemble systems and for some baselines.

We can see that the best-performing feature group, both in terms of accuracy and MAP (65.46 and 83.97, respectively), is the one looking for intra-forum evidence based on search for similar answers in Qatar Living.
It is closely followed by the feature group looking for external evidence in Qatar-related web sites, excluding Qatar Living, which achieved accuracy of 63.45, and the best overall F$_1$ score of 71.65. 

Evidence from high-quality posts in Qatar Living ranks 4th with accuracy of 60.24,
and support from the current thread only comes 7th with accuracy of just 53.41. These results show the importance of forum-level evidence that goes beyond the target thread and beyond known high-quality posts in the forum.

Answer-related features are the third most important feature family. In particular,  linguistic features rank third overall with accuracy of 60.64; this should not be surprising as such features have been shown to be important in previous work~\cite{popat2016credibility}.
We can also see the strong performance of using knowledge about the domain in the form of word embeddings trained on Qatar Living, which are ranked 5th with accuracy of 59.44. However, general word embeddings, e.g., those trained on Google News, do not work well: with accuracy of 52.61, they are barely above the majority class baseline, which has an accuracy of 51.41.

The answer content feature family also contains a group of features that have been previously proposed for modeling credibility. This group achieves an accuracy of 56.23, and we also use it as one of the baselines in the bottom of the table. There are two reasons for its modest performance: \Ni credibility is different from veracity as the former is subjective while the latter is not, and \Nii these features are generally not strong enough by themselves, as they have been originally proposed to work together with features modeling the user (age, followers, friends, etc.), a target topic, and propagation (spreading tree) on Twitter~\cite{Castillo:2011:ICT:1963405.1963500}.

Interestingly, the feature types about the user profile perform the worst. They are also below the majority class baseline in terms of accuracy; however, they outperform the baselines in terms of MAP\@. We believe that the poor performance is due to modeling a user based on her activity, posting categories, and goodness (whether she tries to answer the question irrespective of the veracity of the given answer) of her posts in the past, which do not target factuality directly. In future work, 
we could run our factuality classifier over all of Qatar Living, and we can then characterize a user based on our predicted veracity of his/her answers.

The bottom of the table shows the results for two ensemble systems that combine the above feature groups, yielding accuracy of 72.29 (19 points of improvement over the majority class baseline, absolute) and MAP of 86.54 (23 points of improvement over the chronological baseline, absolute). These results 
indicate that our system might already be usable in real applications.

\section{Discussion}


\begin{table*}[tbh]\footnotesize
\centering
\begin{tabular}{lcccl}
\hline
\multicolumn{5}{l}{\textbf{Q:} does anyone know if there is a french speaking nursery in doha?}\\
\multicolumn{5}{l}{\textbf{A:} there is a french school here. don't know the ages but my neighbor's 3 yr old goes there...}\\
\hline
\multicolumn{5}{l}{\textbf{Best Matched Sentence for Q\&A:} there is a french school here.}\\
\hline
\hline
\textbf{Post Id} & \textbf{sId} & \textbf{R1} & \textbf{R2} & \textbf{Sentence}\\
\hline
35639076 & 15 & 1 & 10 & the pre-school follows the english program but also gives french and arabic lessons.\\ 
32448901 & 4 & 2 & 11 & france bought the property in 1952 and since 1981 it has been home to the french institute.\\ 
31704366 & 7 & 3 & 1 & they include one indian school, two french, seven following the british curriculum...\\
27971261 & 6 & 4 & 4 & the new schools include six qatari, four indian, two british, two american and a finnish...\\
\hline
\end{tabular}
\caption{Sample of sentences from high-quality posts automatically extracted to support the answer \textit{A} to question \emph{Q}. \textit{sId} is the sentence id in the post, \textit{R1} is the ranking based on entailment, and \textit{R2} is the similarity ranking.}
\label{table:article_supports}
\end{table*}

\begin{table*}[tbh]\footnotesize
\centering
\begin{tabular}{@{}l@{}c@{}c p{10.3cm} l@{}}
\hline
\multicolumn{4}{l}{\textbf{Q:} Hi; Just wanted to confirm Qatar's National Day. Is it 18th of December? Thanks.}\\
\multicolumn{4}{l}{\textbf{A:} yes; it is 18th Dec.}\\
\hline
\multicolumn{4}{l}{\textbf{Query generated from Q\&A:} \texttt{"National Day" "Qatar" National December Day confirm wanted}}\\
\hline
\hline
& \textbf{Qatar-} & \textbf{Source} &\\
\textbf{URL} & \textbf{related?} & \textbf{type} & \textbf{Snippet}\\
\hline
\url{qppstudio.net} & No & Other & Public holidays and national ... the world's source of Public holidays information\\ 
\url{dohanews.co} & Yes & Reputed & culture and more in and around Qatar ... The documentary features human interest pieces that incorporate the day-to-day lives of Qatar residents\\ 
\url{iloveqatar.net} & Yes  & Forum & Qatar National Day - Short Info ... the date of December 18 is celebrated each year as the National Day of Qatar...\\
\url{cnn.com} & No & Reputed & The 2022 World Cup final in Qatar will be held on December 18 ... Qatar will be held on December 18 -- the Gulf state's national day.  Confirm. U.S ...\\
\url{icassociation.co.uk} & No & Other & In partnership with ProEvent Qatar, ICA can confirm that the World Stars \\ 
 & & & will be led on the 17 December, World Stars vs Qatar Stars - Qatar National Day.\\
\hline
\end{tabular}
\caption{Sample snippets returned by a search engine for a given query generated from a Q\&A pair. }
\label{table:ir_example}
\end{table*}

\paragraph{High-Quality~Posts}

As explained above, we use a three-step approach to extract supporting evidence from the high-quality posts, namely query generation (Step 1), evidence retrieval using vector-based similarity (Step 2), and re-ranking based on entailment (Step 3). We conducted an ablation experiment in order to investigate the individual contribution of steps 1 and 3. We considered the following settings:

\begin{description}\footnotesize
\item[$S1$:] \emph{Full system}. All features from the three steps are used. 
\item[$S2$:] \emph{No re-ranking}. Only steps 1 and 2 are applied.
\item[$S3$:] \emph{No query generation}. The entire answer is used to extract evidence instead of using the generated query, i.e., only steps 2 and 3 are applied.
\item[$S4$:] \emph{No query generation and no re-ranking}. Only step 2 is applied. As in $S3$, the entire answer is used to retrieve evidence.
\end{description}

The results confirmed \Ni \emph{the importance of generating a good query}: discarding step 1 yields sizable drop in performance by 12 accuracy points when comparing $S4$ to $S2$, and by 4 accuracy points when comparing $S3$ to $S1$; and 
\Nii \emph{the importance of re-ranking based on textual entailment}: discarding step 3 yields 11 accuracy points decrease in performance when comparing $S4$ to $S3$, and 3 accuracy points when comparing $S2$ to $S1$.\footnote{More detailed results are omitted for the sake of brevity.} 

\noindent Table~\ref{table:article_supports} illustrates the effect of the entailment-based re-ranking (step 3). It shows a question ($Q$), an answer to verify ($A$), and the top-4 supporting sentences retrieved by our system, sorted according to the entailment-based re-ranking scores ($R1$). Column $R2$ shows the ranking for the same sentences using vector-based similarity (i.e., without applying step 3). 
We can see that using re-ranking yields better results. For example, the first piece of support in $R1$'s ranking is the best overall, while the same sentence is ranked 10th by $R2$. 
Moreover, the top-ranked evidence in $R2$, although clearly pertinent, is not better than the best one in $R1$.

\paragraph{Linguistic Bias}

We further investigated the effectiveness of the linguistic features. The experimental results show that the top-5 linguistic features are (in this order) \textit{strong subjectivity cues}, \textit{implicatives}, \textit{modals}, \textit{negatives}, and \textit{assertives}. 


\paragraph{External Sources Features}
The query generated from the question--answer pair provides enough context for a quality Web search. The results returned by the search engine are mostly relevant, which indicates that the query generation works well. 
More importantly, as Table~\ref{table:ir_example} shows, the results returned by the search engine are relevant with respect to both the query and the question--answer pair. Note also that, as expected, the results that are Qatar-related and also from a reputed or a forum source tend to be generally more relevant.

\section{Conclusion and Future Work}

We have explored a new dimension in the context of community question answering, which has been ignored so far: checking the veracity of forum answers.
As this is a new problem we created CQA-QL-2016-fact, a specialized dataset which we are releasing freely to the research community.
We further proposed a novel multi-faceted model,
which captures information from the answer content (\emph{what is said and how}), from the author profile (\emph{who says it}), from the rest of the community forum (\emph{where it is said}), and from external authoritative sources of information (\emph{external support}). The evaluation results have shown very strong performance.

In future work, we plan to extend our dataset with additional examples. We would also like to try distant supervision based on known facts, e.g., from high-quality posts, which would allow us to use more training data, thus enabling more sophisticated learning architectures, e.g., based on deep learning.
We also want to improve user modeling, e.g., by predicting factuality for the user's answers and then building a user profile based on that. Finally, we want to explore the possibility of providing justifications for the verified answers and to integrate our system in a real application.

\section*{Acknowledgments}

This research is developed by the Arabic Language Technologies (ALT) group at Qatar Computing Research, HBKU in collaboration with MIT-CSAIL\@. It is part of the Interactive sYstems for Answer Search (Iyas) project.

\fontsize{9.5pt}{10.5pt}\selectfont
\bibliography{bibliography}

\begin{thebibliography}{}

\bibitem[\protect\citeauthoryear{Agichtein \bgroup et al\mbox.\egroup
  }{2008}]{Agichtein:2008:FHC:1341531.1341557}
Agichtein, E.; Castillo, C.; Donato, D.; Gionis, A.; and Mishne, G.
\newblock 2008.
\newblock Finding high-quality content in social media.
\newblock In {\em Proceedings of the International Conference on Web Search and
  Data Mining},  183--194.

\bibitem[\protect\citeauthoryear{Banerjee and
  Han}{2009}]{banerjee-han:2009:NAACLHLT09-Short}
Banerjee, P., and Han, H.
\newblock 2009.
\newblock Answer credibility: A language modeling approach to answer
  validation.
\newblock In {\em Proceedings of the Annual Conference of the North American
  Chapter of the Association for Computational Linguistics},  157--160.

\bibitem[\protect\citeauthoryear{Barr\'{o}n-Cede\~{n}o \bgroup et
  al\mbox.\egroup }{2015}]{barroncedeno-EtAl:2015:ACL-IJCNLP}
Barr\'{o}n-Cede\~{n}o, A.; Filice, S.; Da~San~Martino, G.; Joty, S.;
  M\`{a}rquez, L.; Nakov, P.; and Moschitti, A.
\newblock 2015.
\newblock Thread-level information for comment classification in community
  question answering.
\newblock In {\em Proceedings of the 53rd Annual Meeting of the Association for
  Computational Linguistics and the 7th International Joint Conference on
  Natural Language Processing},  687--693.

\bibitem[\protect\citeauthoryear{Canini, Suh, and Pirolli}{2011}]{Canini:2011}
Canini, K.~R.; Suh, B.; and Pirolli, P.~L.
\newblock 2011.
\newblock Finding credible information sources in social networks based on
  content and social structure.
\newblock In {\em Proceedings of the IEEE Third International Conference on
  Privacy, Security, Risk and Trust and the IEEE Third International Conference
  on Social Computing},  1--8.

\bibitem[\protect\citeauthoryear{Castillo, Mendoza, and
  Poblete}{2011}]{Castillo:2011:ICT:1963405.1963500}
Castillo, C.; Mendoza, M.; and Poblete, B.
\newblock 2011.
\newblock Information credibility on {T}witter.
\newblock In {\em Proceedings of the 20th International Conference on World
  Wide Web},  675--684.

\bibitem[\protect\citeauthoryear{Gencheva \bgroup et al\mbox.\egroup
  }{2017}]{RANLP2017:debates}
Gencheva, P.; Nakov, P.; M\`{a}rquez, L.; Barr\'on-Cede{\~n}o, A.; and Koychev,
  I.
\newblock 2017.
\newblock A context-aware approach for detecting worth-checking claims in
  political debates.
\newblock In {\em Proceedings of the International Conference on Recent
  Advances in Natural Language Processing},  267--276.

\bibitem[\protect\citeauthoryear{Hooper}{1974}]{hooper1974assertive}
Hooper, J.
\newblock 1974.
\newblock {\em On Assertive Predicates}.
\newblock Indiana University Linguistics Club.

\bibitem[\protect\citeauthoryear{Hyland}{2005}]{hyland2005metadiscourse}
Hyland, K.
\newblock 2005.
\newblock {\em Metadiscourse: Exploring Interaction in Writing}.
\newblock Continuum Discourse. Bloomsbury Publishing.

\bibitem[\protect\citeauthoryear{Ishikawa, Sakai, and
  Kando}{2010}]{Ishikawa:10}
Ishikawa, D.; Sakai, T.; and Kando, N.
\newblock 2010.
\newblock Overview of the {NTCIR-8} community {QA} pilot task (part i): The
  test collection and the task.
\newblock In {\em Proceedings of NTCIR-8 Workshop Meeting},  421--432.

\bibitem[\protect\citeauthoryear{Jeon \bgroup et al\mbox.\egroup
  }{2006}]{Jeon:2006:FPQ:1148170.1148212}
Jeon, J.; Croft, W.~B.; Lee, J.~H.; and Park, S.
\newblock 2006.
\newblock A framework to predict the quality of answers with non-textual
  features.
\newblock In {\em Proceedings of the 29th Annual International ACM SIGIR
  Conference on Research and Development in Information Retrieval},  228--235.

\bibitem[\protect\citeauthoryear{Joachims}{1999}]{Joachims:1999:MLS:299094.299104}
Joachims, T.
\newblock 1999.
\newblock Making large-scale support vector machine learning practical.
\newblock In Sch\"{o}lkopf, B.; Burges, C. J.~C.; and Smola, A.~J., eds., {\em
  Advances in Kernel Methods}.
\newblock  169--184.

\bibitem[\protect\citeauthoryear{Jurczyk and
  Agichtein}{2007}]{Jurczyk:2007:DAQ:1321440.1321575}
Jurczyk, P., and Agichtein, E.
\newblock 2007.
\newblock Discovering authorities in question answer communities by using link
  analysis.
\newblock In {\em Proceedings of the Sixteenth ACM Conference on Conference on
  Information and Knowledge Management},  919--922.

\bibitem[\protect\citeauthoryear{Karadzhov \bgroup et al\mbox.\egroup
  }{2017}]{RANLP2017:factchecking:external}
Karadzhov, G.; Nakov, P.; M\`{a}rquez, L.; Barr\'on-Cede{\~n}o, A.; and
  Koychev, I.
\newblock 2017.
\newblock Fully automated fact checking using external sources.
\newblock In {\em Proceedings of the International Conference on Recent
  Advances in Natural Language Processing},  344--353.

\bibitem[\protect\citeauthoryear{Karttunen}{1971}]{karttunen1971Implicatives}
Karttunen, L.
\newblock 1971.
\newblock Implicative verbs.
\newblock {\em Language} 47(2):340--358.

\bibitem[\protect\citeauthoryear{Kouylekov and Negri}{2010}]{Kouylekov:2010}
Kouylekov, M., and Negri, M.
\newblock 2010.
\newblock An open-source package for recognizing textual entailment.
\newblock In {\em Proceedings of the ACL 2010 System Demonstrations},  42--47.

\bibitem[\protect\citeauthoryear{Lita \bgroup et al\mbox.\egroup
  }{2005}]{lita2005qualitative}
Lita, L.~V.; Schlaikjer, A.~H.; Hong, W.; and Nyberg, E.
\newblock 2005.
\newblock Qualitative dimensions in question answering: Extending the
  definitional {QA} task.
\newblock In {\em Proceedings of the AAAI Conference on Artificial
  Intelligence},  1616--1617.

\bibitem[\protect\citeauthoryear{Liu, Hu, and
  Cheng}{2005}]{Liu:2005:OOA:1060745.1060797}
Liu, B.; Hu, M.; and Cheng, J.
\newblock 2005.
\newblock Opinion observer: Analyzing and comparing opinions on the web.
\newblock In {\em Proceedings of the 14th International Conference on World
  Wide Web},  342--351.

\bibitem[\protect\citeauthoryear{Lukasik, Cohn, and
  Bontcheva}{2015}]{lukasik-cohn-bontcheva:2015:ACL-IJCNLP}
Lukasik, M.; Cohn, T.; and Bontcheva, K.
\newblock 2015.
\newblock Point process modelling of rumour dynamics in social media.
\newblock In {\em Proceedings of the 53rd Annual Meeting of the Association for
  Computational Linguistics and the 7th International Joint Conference on
  Natural Language Processing},  518--523.

\bibitem[\protect\citeauthoryear{Lyon, Malcolm, and
  Dickerson}{2001}]{lyon2001detecting}
Lyon, C.; Malcolm, J.; and Dickerson, B.
\newblock 2001.
\newblock Detecting short passages of similar text in large document
  collections.
\newblock In {\em Proceedings of the Conference on Empirical Methods in Natural
  Language Processing},  118--125.

\bibitem[\protect\citeauthoryear{Ma \bgroup et al\mbox.\egroup
  }{2015}]{Ma:2015:DRU}
Ma, J.; Gao, W.; Wei, Z.; Lu, Y.; and Wong, K.-F.
\newblock 2015.
\newblock Detect rumors using time noseries of social context information on
  microblogging websites.
\newblock In {\em Proceedings of the 24th ACM International on Conference on
  Information and Knowledge Management},  1751--1754.

\bibitem[\protect\citeauthoryear{Mihaylov and
  Nakov}{2016}]{SemEval2016:task3:SemanticZ}
Mihaylov, T., and Nakov, P.
\newblock 2016.
\newblock {SemanticZ at SemEval-2016 Task 3}: Ranking relevant answers in
  community question answering using semantic similarity based on fine-tuned
  word embeddings.
\newblock In {\em Proceedings of the 10th International Workshop on Semantic
  Evaluation},  879--886.

\bibitem[\protect\citeauthoryear{Mikolov, Yih, and
  Zweig}{2013}]{mikolov-yih-zweig:2013:NAACL-HLT}
Mikolov, T.; Yih, W.-t.; and Zweig, G.
\newblock 2013.
\newblock Linguistic regularities in continuous space word representations.
\newblock In {\em Proceedings of the 2013 Conference of the North American
  Chapter of the Association for Computational Linguistics: Human Language
  Technologies},  746--751.

\bibitem[\protect\citeauthoryear{Morris \bgroup et al\mbox.\egroup
  }{2012}]{Morris:2012:TBU:2145204.2145274}
Morris, M.~R.; Counts, S.; Roseway, A.; Hoff, A.; and Schwarz, J.
\newblock 2012.
\newblock Tweeting is believing?: Understanding microblog credibility
  perceptions.
\newblock In {\em Proceedings of the ACM 2012 Conference on Computer Supported
  Cooperative Work},  441--450.

\bibitem[\protect\citeauthoryear{Mukherjee and
  Weikum}{2015}]{mukherjee2015leveraging}
Mukherjee, S., and Weikum, G.
\newblock 2015.
\newblock Leveraging joint interactions for credibility analysis in news
  communities.
\newblock In {\em Proceedings of the 24th ACM International Conference on
  Information and Knowledge Management},  353--362.

\bibitem[\protect\citeauthoryear{Nakov \bgroup et al\mbox.\egroup
  }{2015}]{nakov-EtAl:2015:SemEval}
Nakov, P.; M\`{a}rquez, L.; Magdy, W.; Moschitti, A.; Glass, J.; and Randeree,
  B.
\newblock 2015.
\newblock {SemEval}-2015 task 3: Answer selection in community question
  answering.
\newblock In {\em Proceedings 9th International Workshop on Semantic
  Evaluation},  269--281.

\bibitem[\protect\citeauthoryear{Nakov \bgroup et al\mbox.\egroup
  }{2016}]{nakov-EtAl:2016:SemEval}
Nakov, P.; M\`{a}rquez, L.; Moschitti, A.; Magdy, W.; Mubarak, H.; Freihat,
  A.~A.; Glass, J.; and Randeree, B.
\newblock 2016.
\newblock {SemEval}-2016 task 3: Community question answering.
\newblock In {\em Proceedings of the 10th International Workshop on Semantic
  Evaluation},  525--545.

\bibitem[\protect\citeauthoryear{Nakov \bgroup et al\mbox.\egroup
  }{2017a}]{nakov-EtAl:2017:SemEval}
Nakov, P.; Hoogeveen, D.; M\`{a}rquez, L.; Moschitti, A.; Mubarak, H.; Baldwin,
  T.; and Verspoor, K.
\newblock 2017a.
\newblock {SemEval-2017} task 3: Community question answering.
\newblock In {\em Proceedings of the 11th International Workshop on Semantic
  Evaluation},  27--48.

\bibitem[\protect\citeauthoryear{Nakov \bgroup et al\mbox.\egroup
  }{2017b}]{RANLP2017:credibility:trolls}
Nakov, P.; Mihaylova, T.; M\`arquez, L.; Shiroya, Y.; and Koychev, I.
\newblock 2017b.
\newblock Do not trust the trolls: Predicting credibility in community question
  answering forums.
\newblock In {\em Proceedings of the International Conference on Recent
  Advances in Natural Language Processing},  551--560.

\bibitem[\protect\citeauthoryear{Pelleg \bgroup et al\mbox.\egroup
  }{2016}]{Pelleg:2016:CEI:2818048.2820022}
Pelleg, D.; Rokhlenko, O.; Szpektor, I.; Agichtein, E.; and Guy, I.
\newblock 2016.
\newblock When the crowd is not enough: Improving user experience with social
  media through automatic quality analysis.
\newblock In {\em Proceedings of the 19th ACM Conference on Computer-Supported
  Cooperative Work \& Social Computing},  1080--1090.

\bibitem[\protect\citeauthoryear{Popat \bgroup et al\mbox.\egroup
  }{2016}]{popat2016credibility}
Popat, K.; Mukherjee, S.; Str{\"o}tgen, J.; and Weikum, G.
\newblock 2016.
\newblock Credibility assessment of textual claims on the web.
\newblock In {\em Proceedings of the 25th ACM International on Conference on
  Information and Knowledge Management},  2173--2178.

\bibitem[\protect\citeauthoryear{Potthast \bgroup et al\mbox.\egroup
  }{2013}]{potthast2013overview}
Potthast, M.; Hagen, M.; Gollub, T.; Tippmann, M.; Kiesel, J.; Rosso, P.;
  Stamatatos, E.; and Stein, B.
\newblock 2013.
\newblock Overview of the 5th international competition on plagiarism
  detection.
\newblock In {\em Proceedings of the CLEF Conference on Multilingual and
  Multimodal Information Access Evaluation},  301--331.

\bibitem[\protect\citeauthoryear{Recasens, Danescu-Niculescu-Mizil, and
  Jurafsky}{2013}]{Recasens:ACL:13}
Recasens, M.; Danescu-Niculescu-Mizil, C.; and Jurafsky, D.
\newblock 2013.
\newblock Linguistic models for analyzing and detecting biased language.
\newblock In {\em Proceedings of the 51st Annual Meeting of the Association for
  Computational Linguistics},  1650--1659.

\bibitem[\protect\citeauthoryear{Riloff and
  Wiebe}{2003}]{Riloff:2003:LEP:1119355.1119369}
Riloff, E., and Wiebe, J.
\newblock 2003.
\newblock Learning extraction patterns for subjective expressions.
\newblock In {\em Proceedings of the Conference on Empirical Methods in Natural
  Language Processing},  105--112.

\bibitem[\protect\citeauthoryear{Su, yun Chen, and
  Huang}{2010}]{Su-EtAl:2010:PACLIC2010}
Su, Q.; yun Chen, H.~K.; and Huang, C.-R.
\newblock 2010.
\newblock Incorporate credibility into context for the best social media
  answers.
\newblock In {\em Proceedings of the 24th Pacific Asia Conference on Language,
  Information and Computation},  535--541.

\bibitem[\protect\citeauthoryear{Zubiaga \bgroup et al\mbox.\egroup
  }{2016}]{PlosONE:2016}
Zubiaga, A.; Liakata, M.; Procter, R.; Wong Sak~Hoi, G.; and Tolmie, P.
\newblock 2016.
\newblock Analysing how people orient to and spread rumours in social media by
  looking at conversational threads.
\newblock {\em PLoS ONE} 11(3):1--29.

\end{thebibliography}
\bibliographystyle{aaai}

\end{document}